\documentclass{article}

\usepackage{microtype}
\usepackage{graphicx}
\usepackage{subfigure}
\usepackage{booktabs} 
\usepackage{multirow}
\usepackage{colortbl}
\usepackage{rotating}
\usepackage{enumitem}
\usepackage{lipsum}
\usepackage{subcaption}
\usepackage[table,xcdraw]{xcolor}
\newcommand{\thickhline}{\noalign{\hrule height 1pt}}
\usepackage{hyperref}

\usepackage[accepted]{icml2024}

\usepackage{amsmath}
\usepackage{amssymb}
\usepackage{mathtools}
\usepackage{amsthm}

\usepackage[capitalize,noabbrev]{cleveref}

\theoremstyle{plain}

\theoremstyle{definition}

\theoremstyle{remark}

\usepackage[textsize=tiny]{todonotes}

\def\ie{\emph{i.e.,~}}

\icmltitlerunning{3D-VLA: A 3D Vision-Language-Action Generative World Model}
\usepackage{xspace}
\begin{document}
\newcommand{\modelname}[0]{3D-VLA\xspace}
\twocolumn[
    \icmltitle{3D-VLA: A 3D Vision-Language-Action Generative World Model}

    \icmlsetsymbol{equal}{*}
    \begin{icmlauthorlist}
\icmlauthor{Haoyu Zhen}{umass,sjtu}
\icmlauthor{Xiaowen Qiu}{umass}
\icmlauthor{Peihao Chen}{scut}
\icmlauthor{Jincheng Yang}{sjtu}
\icmlauthor{Xin Yan}{whu}\\
\icmlauthor{Yilun Du}{mit}
\icmlauthor{Yining Hong}{ucla}
\icmlauthor{Chuang Gan}{umass,mit-ibm}
\end{icmlauthorlist}

\icmlaffiliation{umass}{University of Massachusetts Amherst}
\icmlaffiliation{sjtu}{Shanghai Jiao Tong University}
\icmlaffiliation{scut}{South China University of Technology}
\icmlaffiliation{whu}{Wuhan University}
\icmlaffiliation{ucla}{University of California, Los Angeles
}
\icmlaffiliation{mit}{Massachusetts Institute of Technology}
\icmlaffiliation{mit-ibm}{MIT-IBM Watson AI Lab}

    \begin{center}{
        \hypersetup{urlcolor=red}
        \url{https://vis-www.cs.umass.edu/3dvla}
    }\end{center}
    \icmlkeywords{Machine Learning, ICML}
    \vskip 0.3in
]



\printAffiliationsAndNoticeArxiv{}  

\begin{abstract}
Recent vision-language-action (VLA) models rely on 2D inputs, lacking integration with the broader realm of the 3D physical world. Furthermore, they perform action prediction by learning a direct mapping from perception to action, neglecting the vast dynamics of the world and the relations between actions and dynamics. In contrast, human beings are endowed with world models that depict imagination about future scenarios to plan actions accordingly.
To this end, we propose \modelname by introducing a new family of embodied foundation models that seamlessly link 3D perception, reasoning, and action through a generative world model. Specifically, \modelname is built on top of a 3D-based large language model (LLM), and a set of interaction tokens is introduced to engage with the embodied environment.
Furthermore, to inject generation abilities into the model, we train a series of embodied diffusion models and align them into the LLM for predicting the goal images and point clouds.
To train our 3D-VLA, we curate a large-scale 3D embodied instruction dataset by extracting vast 3D-related information from existing robotics datasets. Our experiments on held-in datasets demonstrate that \modelname significantly improves the reasoning, multimodal generation, and planning capabilities in embodied environments, showcasing its potential in real-world applications.
\end{abstract}

\section{Introduction}
\label{sec:intro}

Nowadays, there has been a proliferation of vision-language models~\cite{liu2023visual, alayrac2022flamingo, li2023blip} that can take images as inputs and perform a series of reasoning tasks in the 2D space, mirroring the versatility of the human brain.
Such 2D foundation models also lay the foundation for recent embodied foundation models such as RT-2~\cite{brohan2023rt} and PALM-E~\cite{driess2023palm} that could generate high-level plans or low-level actions contingent on the images.
However, they neglect the fact that human beings are situated within a far richer 3D physical world beyond 2D images -  they reason, plan, and act based on their 3D understanding of the environment \cite{effects, seeing, 10.7551/mitpress/9780262514620.001.0001}. It's crucial that human-like intelligent embodied agents are equipped with the same 3D understanding ability.

Taking a step forward, recent works~\cite{huang2023embodied, hong2024multiply} develop embodied foundation models that could plan and act in the 3D environment.
However, such models mainly learn a direct mapping from perception to action, devoid of a broader understanding of the dynamics of the world, and the relations between actions and world dynamics.
On the other hand, human beings are blessed with world models that simulate future events based on 3D internal representations.
By depicting the imagination and anticipation about the future states, one could better plan actions toward the predicted goals.

Challenges inevitably exist for building such human-like 3D world models.
Firstly, existing foundation models focus on language generation, unable to imagine modalities beyond language and simulate future states to facilitate action generation, which is a crucial aspect of world models. Secondly, existing embodied datasets mainly contain 2D images or videos, lacking 3D-related annotations for reasoning and planning in the 3D space.

\begin{figure*}[t]
    \centering
    \includegraphics[width=0.95\linewidth]{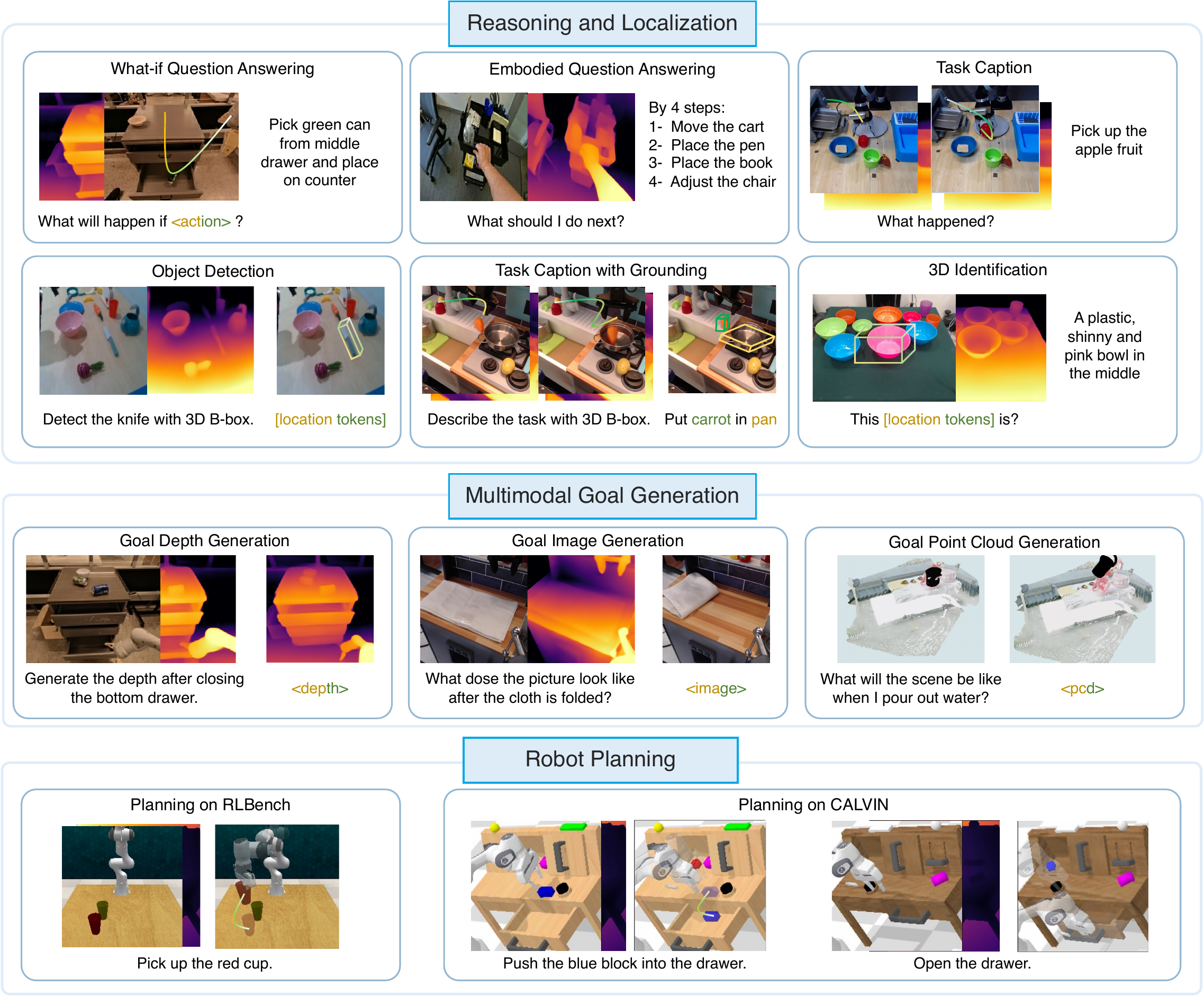}
    \vspace{-0.4cm}
    \caption{Examples from our 3D Embodied Instruction Tuning Dataset.}
    \vspace{-0.4cm}

    \label{fig:enter-label}
\end{figure*}

To this end, we propose \modelname by introducing a new family of embodied foundation models that seamlessly link 3D perception, reasoning, and action through a generative world model.
Specifically, we build our \modelname on top of a 3D large language model~\cite{hong20233d} to equip the model with 3D understanding ability.
Since embodied tasks could not be accomplished via language generation solely and require deeper digging into the dynamic scenes, the manipulated objects as well as actions to interact with the scenes,
we add special interactive tokens to the LLM vocabulary (\textit{e.g.,} scene, object, and action tokens). These added tokens enable our model to perform a wider range of embodied tasks and support interleaved 3D-text data.
Recognizing the inadequacy of multimodal generation ability in embodied foundation models, we propose to inject the goal generation ability into \modelname.
We first pretrain a set of embodied diffusion models for RGBD-to-RGBD and point-to-point generation respectively.
To efficiently bridge between the diffusion decoders of various modalities and the LLM embedding space,
we employ a projector that aligns multi-modal goal generation in \modelname.
It strategically incorporates multimodal signals to specify the type of modality for a generation.

Another challenge for building such a generative world model lies in the lack of data.
The embodied datasets in use~\cite{padalkar2023open, brohan2022rt, jang2022bc} mainly consist of 2D images,
deficient in 3D-related information.
Thus, we curate a large-scale 3D embodied instruction tuning dataset. Specifically, we first gather a diverse collection of datasets that includes real and synthetic data featuring robot manipulations and human-object interactions.
For datasets lacking depth data, we utilize a depth estimator to append necessary 3D details and project them to 3D point clouds. Additionally, we design a pipeline to use the off-the-shelf models to extract 3D-related annotations and enrich the language descriptions. In this way, we collect 2M 3D-language-action data pairs, covering various tasks such as task captioning, action prediction, localization, multimodal goal generation, etc, as shown in Figure \ref{fig:enter-label}.

\begin{figure*}[th]
    \centering
    \includegraphics[width=\linewidth]{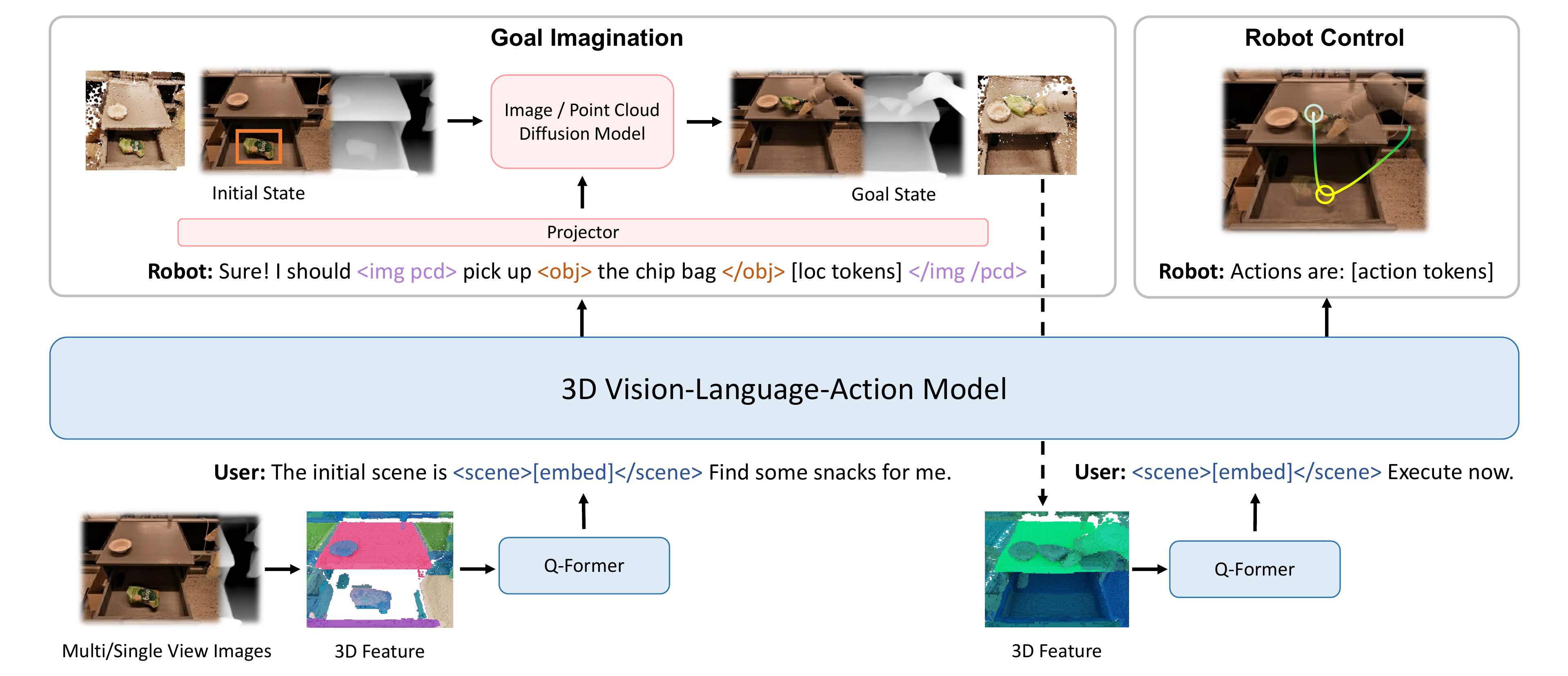}
    \vspace{-5mm}
    \caption{Overview of our \modelname pipeline. The left part shows our goal-generation capability. Our model can imagine the final state image and point cloud based on the user's input. This generated goal state can then be fed back to our model to guide the robot control.}

    \label{fig:method}
\end{figure*}

To sum up, we have the following contributions:
\vspace{-2mm}
\begin{itemize}[align=right,itemindent=0em,labelsep=2pt,labelwidth=1em,leftmargin=*,itemsep=0em]
    \item We propose \modelname, a new family of 3D vision-language-action embodied foundation models that unify 3D perception, reasoning, and action with a generative world model.

    \item We create a large-scale 3D embodied instruction tuning dataset addressing the absence of 3D-related information in existing embodied datasets.
    \item We add interaction tokens to better interact with the environment. We further train diffusion models for goal image and point cloud generation. We utilize a projector to efficiently align LLM output features and diffusion models.
    \item Our \modelname can conduct a series of tasks, including goal generation (in terms of images, depths, and point clouds), goal-based planning, and embodiment action prediction. It outperforms baseline models by a large margin in these novel embodied tasks. It also outshines baseline models in traditional language-based tasks.
\end{itemize}

\section{Related Works}
\textbf{Multimodal Language Models} Recent Multimodal Language Models have made remarkable advances in various domains, including vision and language understanding~\citep{li2022blip, li2023blip, liu2023visual, huang2023language, peng2023kosmos, zhu2023minigpt}, interleaved image and text understanding~\citep{alayrac2022flamingo}, interleaved image and text generation~\citep{dong2023dreamllm}. Some more unified models can perceive inputs and generate outputs in arbitrary combinations of text, images, videos, and audio~\cite{wu2023next, lu2023unifiedio}. However, none of these models can perceive 3D inputs or output actions according to 3D input.

\textbf{Vision-Language-Action Models} Previous vision-language models with action output have predominantly leveraged 2D features, thereby lacking the capability of 3D spatial understanding~\citep{driess2023palme, brohan2022rt, brohan2023rt}. In contrast, our model is guided by 3D features, which are predicted in alignment with goal objectives in our general world model. We are the first to leverage 3D features such as point clouds for action token generation, significantly improving action planning accuracy. 
Additionally, this pipeline possesses the potential to be extended for applications in real-world scenarios.

\textbf{3D Foundation Models} Our paper is closely related to the 3D foundation models that integrate 3D features in MLLMs~\citep{hong20233d, chen2023ll3da, qi2023gpt4point, xu2023pointllm, huang2023chat3d, zhou2023uni3d, guo2023pointbind, li20243dmit}. These studies have successfully stepped forward to leverage foundation models to comprehend 3D features. However, they primarily focus on analyzing and reasoning in the current observable state of the 3D scenes, thereby revealing a limitation in predicting future features that extend beyond immediate perception. Contrasting with them, we aim to not only understand the perceivable scenes but also predict imperceptible multimodal features guided by specific goals. This capability enables our model to further generate action tokens to interact with the 3D world.

\section{3D Embodied Instruction Tuning Dataset}
Recently, benefiting from billion-scale datasets on the internet, VLMs have demonstrated exceptional proficiency in various tasks. 
Similarly, million-level datasets comprising video-action pairs lay the foundation for embodied VLMs for robot control. 
However, they mostly don't provide depth or 3D annotations and precise control in robot operations that necessitate the inclusion of 3D spatial reasoning and interaction. 
Without 3D information, it is challenging for a robot to comprehend and execute the commands that require 3D spatial reasoning, such as ``place the farthest cup into the middle drawer".

To bridge this gap, we build a large-scale 3D embodied instruction tuning dataset that provides sufficient 3D-related information as well as paired text instructions to train our model.
We design a pipeline to extract 3D-language-action pairs from existing embodied datasets,  obtaining annotations for point clouds, depth maps, 3D bounding boxes, the robot's 7D actions, and textual descriptions. The details are outlined as follows.

\subsection{Dataset Collection} 

Our data are curated from various sources. We provide an overview here, with details available in the Appendix:

\textbf{Robot Datasets:} We select 12 datasets~\cite{brohan2022rt, jang2022bc, walke2023bridgedata, lynch2023interactive, feng2023finetuning, chen2023playfusion, dass2023jacoplay, mandlekar2019scaling, mees23hulc2, shah2023mutex, sawhney2021playing, robovqa2023arxiv} from the Open-X Embodiment Dataset~\cite{padalkar2023open}. They have high-quality images with linguistic instructions in the real world but lack more in-depth information and 3D annotations. We also select datasets with excellent depth information, such as Dobb-E~\cite{shafiullah2023bringing} and RH20T~\cite{fang2023rh20t}. Additionally, we use datasets collected from two simulator environments, RLBench~\cite{james2020rlbench} and CALVIN~\cite{mees2022calvin}.

\textbf{Human Object Interaction Datasets}: Human/hand-object interactions could provide demonstrations that benefit
 robot decision-making and imitation. Therefore, we utilize several human-object interaction datasets, including datasets without depth information, such as Epic-Kitchens~\cite{damen2018scaling}, and datasets with better 3D annotations, such as HOI4D~\cite{liu2022hoi4d}.

\subsection{Visual Annotations}

\textbf{Estimating depths and optical flows.}
Given that over 95\% of the video datasets for embodied tasks do not provide 3D information, we employ ZoeDepth~\cite{bhat2023zoedepth} on each frame of the video from these datasets. Additionally, to better utilize video data, we use RAFT~\cite{teed2020raft} for optical flow estimation. Optical flow aids in refining the data we generate. Thus, for video segments where the camera pose does not change, we use optical flow to estimate which pixels are the unmoved background. We align the depth maps of these backgrounds across different frames of the same video, multiplying each frame's depth map by a coefficient to ensure depth consistency. After getting the depth maps, we can directly lift the RGB-D images into 3D point clouds using camera intrinsics and poses.

\textbf{Generating 3D annotations.}
\noindent We aim to generate several 3D-related annotations: 3D bounding boxes of the objects, goal images, depths, or point clouds as the imagination outcomes, as well as robot actions in the 3D space. 
 We first extract the 3D bounding boxes of the objects in the scenes. Such information could benefit 3D models' ability to capture 3D information and attend to the manipulated object for better decision-making. The embodied datasets that serve as sources provide text instructions to describe the commands executed by the robots. We use spaCy~\cite{spacy2} to parse the instructions to obtain all noun chunks, including the manipulated object.
We utilize a pre-trained grounding model (e.g., Grounded-SAM~\cite{ren2024grounded} ) to obtain the 2D mask of each object. These 2D masks, when lifted to 3D,  correspond to parts of the point cloud, allowing us to obtain the 3D bounding boxes of all the objects in space. 
When selecting masks, the manipulated object is chosen based on the highest confidence value in areas of significant optical flow. 
Since we reconstruct the depths and point clouds, we could use images, depths, and point clouds in future frames as ground-truth goals. For actions, we use the 7 DoF actions from the provided datasets.

\subsection{Language Annotations}
Inspired by~\citep{li2023covlm, peng2023kosmos}, we propose to generate dense language annotations consisting of tokens (\textit{e.g.,} \texttt{\small<image></image>}; \texttt{<pcd></pcd>}) that encompass the 3D annotations (bounding boxes, goal images / depths / point clouds, actions) we generated before, as shown in the prompts in Figure \ref{fig:method}. 

\noindent We use pre-defined language templates with tokens to construct these 3D annotations into prompts and answers. Following~\cite{hong20233d}, we use ChatGPT-based prompting to diversify prompts. Specifically, we provide instructions to ChatGPT, as well as our annotated objects and bounding boxes. We also give 2-3 few-shot human-written demonstrations to guide the GPT on the type of data it is instructed to generate. ChatGPT is asked to summarize the information and rewrite the template-generated prompts into more diverse forms. For tasks without pre-defined templates, ChatGPT is also asked to generate prompts and answers as language inputs and outputs of these tasks by itself. 
We show the detailed templates and prompts to generate all types of data in the Appendix.

\section{Methods}
\subsection{Overview}

In this section, we introduce \modelname, a world model for 3D  reasoning, goal generation, and decision-making in embodied environments. As shown in Figure \ref{fig:method}, we first build our backbone on top of 3D-LLM \cite{hong20233d}, and further enhance the model's capabilities to interact with the  3D world by adding a series of interaction tokens. Next, we inject goal generation ability into \modelname by first pretraining the embodied diffusion models and employing a projector for aligning the LLM and the diffusion models.


\subsection{3D-VLA}
\subsubsection{Backbone}
In the first stage, we develop the \modelname base model following the methodology of 3D-LLM \cite{hong20233d}. 
Since the dataset we collected is not at the billion-level scale required for training a multi-modal LLM from scratch, we follow the approach of 3D-LLM by leveraging multi-view features to generate 3D scene features. This enables the seamless integration of visual features into a pre-trained VLM with no need for adaptation.
Meanwhile, the training datasets for 3D-LLM mostly comprise objects \cite{deitke2022objaverse} and indoor scenes \cite{dai2017scannet, ramakrishnan2021habitatmatterport}, which do not directly align with our embodied setup. Therefore, we choose not to load the 3D-LLM pretrained model. Instead, we utilize BLIP2-FlanT5\textsubscript{XL}~\cite{li2023blip} as our pretrained model. During training, we unfreeze both the input and output embeddings for tokens, as well as the weights of the Q-Former.

\subsubsection{Interaction Tokens}
To enhance the model's comprehension of 3D scenes and facilitate interaction within these environments, we introduce a novel set of interaction tokens. 
Firstly,  We incorporate object tokens \texttt{\small<obj> </obj>} that enclose the object nouns in the parsed sentences (\textit{e.g.},  \texttt{\small<obj> a chocolate bar </obj> [loc tokens] on the table}) so that the model could better capture which objects are manipulated or referred to. 
Secondly, to better represent spatial information by language, we devise a set of location tokens \texttt{\small<loc0-255>} for grounding referred objects, which are represented by six tokens for the 3D bounding box in the form of AABB. 
Thirdly, to better encode dynamics with our framework, we introduce the \texttt{\small<scene> </scene>} tokens to enclose the embeddings of a static scene. By composing over the scene tokens, \modelname could comprehend dynamic scenes and manage inputs that interleave 3D scenes and text.

\noindent We further enhance the architecture with an expanded set of specialized tokens that represent robotic actions. The robot's actions, with 7 degrees of freedom, are represented by discrete tokens such as \texttt{\small<aloc0-255>}, \texttt{\small<arot0-255>}, and \texttt{\small<gripper0/1>} to denote the arm's intended absolute location, rotation, gripper openness. These actions are separated by token \texttt{\small<ACT\_SEP>}. 

\subsection{Injecting Goal Generation Ability into 3D-VLA}
In this section, we introduce how our \modelname performs goal generation in terms of images, depths, and point clouds.

\noindent Human beings pre-visualize the final states of the scenes to facilitate action prediction or decision making, which is a key aspect in building world models.
Moreover, during preliminary experiments, we also discover that providing the ground-truth final states can enhance the model's reasoning and planning capabilities.
However, training an MLLM to generate images, depths, and point clouds is non-trivial. 
Firstly, state-of-the-art video diffusion models are not tailored for embodied setups. For instance, when asking Runway \cite{esser2023structure} to generate future frames given the instruction ``open the drawer", the entire scene is altered to a great extent with regard to view change, unexpected object deformation, and weird texture replacement, as well as layout distortion. Similarly, using the method of DreamLLM~\cite{dong2023dreamllm} to directly freeze the stable diffusion trained on internet data, can lead to collapsed outputs. 
Secondly, how to incorporate diffusion models of various modalities into a single foundation model remains a challenge.
Therefore, we propose to inject the ability to generate images, depths and point clouds into \modelname. We first pretrain the embodied diffusion models in terms of different modalities such as images, depths and point clouds, and then align the decoders of these diffusion models to the embedding space of 3D-VLA through an alignment stage.

\begin{table*}[t]
\centering
\resizebox{\textwidth}{!}{
\begin{tabular}{cl|lllllll}
\thickhline
\small
 \rule{0pt}{2.6ex}Tasks &
  Models &
  \multicolumn{1}{c}{BLEU-1} &
  \multicolumn{1}{c}{BLEU-2} &
  \multicolumn{1}{c}{BLEU-3} &
  \multicolumn{1}{c}{BLEU-4} &
  \multicolumn{1}{c}{METEOR} &
  \multicolumn{1}{c}{ROUGH-L} &
  \multicolumn{1}{c}{EM@1} \rule[-1.2ex]{0pt}{0pt}\\ \hline
\multirow{6}{*}{Embodied QA} 
& 3D-LLM$^*$                 & 1.05 & 0.38 & 0.15 & 0.02 & 12.96 & 0.91 & 0.00 \\
                             & BLIP2 OPT\textsubscript{2.7B}$^*$     & 7.39 & 3.17 & 0.03 & 0.02 & 3.87 & 7.40 & 3.03 \\
                             & BLIP2 FlanT5\textsubscript{XL}$^*$    & 22.84 & 16.17 & 12.50 & 10.11 & 11.41 & 32.01 & 10.31 \\
                             & OpenFlamingo\textsubscript{4B}$^*$    & 9.50 & 6.51 & 5.14 & 4.29 & 6.84 & 10.40 & 1.21 \\
                             & LLaVA\textsubscript{7B}$^*$           & 11.66 & 8.06 & 6.01 & 4.58 & 12.59 & 14.17 & 5.67 \\
                             & BLIP2 FlanT5\textsubscript{XL}        & 37.31 & 27.20 & 20.32 & 15.48 & 17.80 & 38.92 & 15.35 \\
                             & \textbf{\modelname}                   & \textbf{48.34} & \textbf{38.55} & \textbf{31.72} & \textbf{26.80} & \textbf{23.72} & \textbf{49.33} & \textbf{24.53} \\ \hline
\multirow{6}{*}{Task Caption} 
                             & 3D-LLM$^*$                 & 0.78 & 0.16 & 0.07 & 0.05 & 0.57 & 1.33 & 0.00 \\
                             & BLIP2 FlanT5\textsubscript{XL}$^*$     & 8.50 & 2.07 & 0.35 & 0.00 & 3.40 & 8.45 & 0.00 \\
                             & OpenFlamingo\textsubscript{4B}$^*$     & 7.61 & 1.64 & 0.37 & 0.00 & 4.74 & 9.36 & 0.00 \\
                             & LLaVA\textsubscript{7B}$^*$            & 2.63 & 0.69 & 0.16 & 0.00 & 2.63 & 4.65 & 0.00 \\
                             & BLIP2 FlanT5\textsubscript{XL}         & 22.05 & 11.40 & 5.72 & 3.16 & 8.72 & 26.12 & 7.75 \\
                             & \textbf{\modelname}                   & \textbf{55.69} & \textbf{45.88} & \textbf{39.39} & \textbf{34.88} & \textbf{27.57} & \textbf{62.01} & \textbf{29.34} \\ \hline
\multirow{2}{*}{What-if QA}
                             & BLIP2 FlanT5\textsubscript{XL}     & 28.23 & 11.47 & 4.49 & 0.06 & 8.27 & 28.41 & 5.85 \\
                             & \textbf{\modelname}                   & \textbf{53.09} & \textbf{40.94} & \textbf{34.34} & \textbf{29.38} & \textbf{26.83} & \textbf{52.82} & \textbf{14.7} \\ \hline
\multirow{3}{*}{\begin{tabular}[c]{@{}c@{}}Dense Caption\\ \end{tabular}} 
                             & 3D-LLM$^*$                 & 0.52 & 0.22 & 0.16 & 0.13 & 0.34 & 0.64 & 0.00 \\
                             & BLIP2 FlanT5\textsubscript{XL}     & 36.17 & 24.72 & 18.06 & 13.96 & 17.83 & 40.56 & 13.10 \\
                             & \textbf{\modelname}                   & \textbf{51.90} & \textbf{42.83} & \textbf{38.11} & \textbf{34.62} & \textbf{25.25} & \textbf{55.91} & \textbf{39.49} \\ \hline
\end{tabular}}
\caption{Evaluation on reasoning ability using held-in data. $*$ denotes zero-shot transfer results without training on our pre-train datasets. 
}
\label{tab:pretrain_held_in}
\end{table*}

\subsubsection{Pretraining Embodied Diffusion Models for Goal Generation}
To address the limitations of current diffusion models for goal generation in an embodied environment, we train RGB-D to RGB-D and point-cloud to point-cloud diffusion models. We utilize our curated 3D-language video data to train a conditional diffusion model that edits the initial state modality based on instructions to generate the corresponding final state modality. 
The specific training details for these models are as follows: For RGBD to RGBD generation, we employ Stable Diffusion V1.4~\cite{rombach2022high} as our pretrained model due to the efficiency and quality of image generation by latent diffusion when operating in the latent space of a pretrained VAE~\cite{kingma2013auto}. We concatenate the RGB latent and depth latent as the image condition. Similarly, for point-to-point generation, we use Point-E~\cite{nichol2022point} as the pretrained model, to which we add a point cloud condition input.

\subsubsection{Bridging LLM and Goal Generation}
After pretraining the diffusion models, we are equipped with various decoders that could generate goals by conditioning the latent spaces in their modalities. 
Challenges remain as to how to seamlessly incorporate the pretrained decoders into the LLMs so that \modelname could generate goals with regard to any pretrained modalities conditioned on the input instructions.
To bridge the gap between the LLM and the diffusion models of different modalities, we develop an alignment stage into our \modelname.
We first introduce additional special tokens such as \texttt{\small<image>} \texttt{\small</image>} and \texttt{\small<pcd>} \texttt{\small</pcd>}. These tokens are intricately designed to inform the decoder about the type of modal content to output. 
Between the enclosing tokens, we supervise the LLM in generating instructions for a robot to execute, which may include object tokens and location tokens, such as \texttt{\small<image> pick up the <obj> apple </obj> [loc tokens] </image>}. 
Based on this, we can apply a transformer-based projector, which is capable of mapping the decoder features and embeddings from the Large Language Model (LLM) into the space of the DM framework. It plays a crucial role in enhancing the model's capability to understand and generate multi-modal data, establishing a connection between high-level language understanding and multi-modal goal generation. To make training \modelname more efficient and to avoid catastrophic forgetting, we utilize LoRA~\cite{hu2021lora} to fine-tune different diffusion models. At the same time, we only train the newly introduced special tokens embeddings, the corresponding embedding output linear layer, and the entire projector. We minimize both the LLM and DM denoising loss.

\section{Experiments}
3D-VLA is a versatile 3D-based generative world model that can perform reasoning and grounding in the 3D world, imagine multi-modal goal content, and generate actions for robot manipulation. In this section, we evaluate 3D-VLA in three aspects: 3D reasoning and localization, multi-modal goal generation, and embodied action planning.

\subsection{3D Reasoning and Localization}
\textbf{Tasks.} Our primary focus is on scenes involving robots that are characterized by greater dynamism and a higher degree of interaction, which require a greater level of reasoning and localization abilities. 
We build several tasks on 3D embodied instruction tuning datasets for learning these abilities in the robotics domain. The tasks include 1) embodied QA on RoboVQA dataset~\cite{robovqa2023arxiv}; 2) task captioning on 11 Open-X datasets~\cite{padalkar2023open}, where we input the initial and final scenes and ask the agent to reason what has happened; 3) what-if QA on RT-1 dataset~\cite{brohan2022rt}, where the agent is asked a question that what will happen if some specified actions (represented by action tokens) are executed; 4) dense captioning on 11 Open-X datasets, where the agent need to caption the content specified by a 3d bounding box; 5) localization on 11 Open-X datasets, where the agent is to localize the object mentioned in the robot manipulation instruction.
We evaluate \modelname on these tasks using held-in datasets. 

\begin{table}[t]
\begin{tabular}{l|ccc}
\thickhline
Methods & IoU & Acc@25 & \multicolumn{1}{c}{Acc@50} \\ \hline
Kosmos-2 (w/ GT Depth) & 10.92 & 12.73 & 3.85 \\
CoVLM (w/ GT Depth) & 19.81 & 25.39 & 16.61 \\
\modelname              & \textbf{29.33} & \textbf{42.26} & \textbf{27.09} \\ \thickhline
\end{tabular}
        \vspace{-4mm}

\caption{Localization results on held-in robotics datasets. }
\label{tab:localization}
\end{table}

\begin{table}[t]
\centering
\resizebox{\linewidth}{!}{
\begin{tabular}{l|cccc}
\thickhline
\rule{0pt}{2.6ex}\rule[-1.2ex]{0pt}{0pt}
Method & PSNR $\uparrow$ & CLIP Sim $\uparrow$ & SSIM $\uparrow$ & FID $\downarrow$ \\ \hline
Instruct-P2P     &    14.41       & 0.909 & 0.389 & 0.309 \\
SuSIE            &    15.20       &  0.898 & 0.549 & 0.182 \\ 
NeXT-GPT         &    8.86       &  0.199 & 0.153 & 0.432 \\ 
Instruct-P2P$^*$ & 16.67 & \textbf{0.941} & 0.628 & 0.178 \\
\hline
\modelname w/o Pred BBox           & 17.02          & 0.919 & 0.632 & \textbf{0.173} \\
\modelname              & \textbf{17.21}          & 0.920 & \textbf{0.636} & 0.177 \\ 
\thickhline
\end{tabular}}
\caption{RGB image goal generation results. $*$ denotes the model is trained on our pretrained dataset.
\label{tab:goal_image1}
}
\end{table}

\begin{table}[t]
\centering
\begin{tabular}{l|cc}
\thickhline
Models & P-FID $\downarrow$ & Chamfer-$L_1$ $\downarrow$\\ \hline
Point-E$^*$ & 5.241 & 0.159 \\
\modelname w/o Pred BBox             & 4.914 & 0.143 \\
\modelname       & \textbf{4.796} & \textbf{0.139} \\ \thickhline
\end{tabular}
\vspace{-2mm}

\caption{Point Cloud goal generation results. $*$ denotes the model is trained on our pretrained dataset.}

\label{tab:goal_pcd}
\end{table}

\textbf{Baselines.}
We compare 3D-VLA with 3D-LLM~\cite{hong20233d} and 2D vision-language models, including BLIP2~\cite{li2023blip}, OpenFlamingo~\cite{alayrac2022flamingo}, and LLaVA~\cite{liu2023visual}. We implement these baselines in two ways: 1) zero-shot transfer where we test the released trained model on these new tasks; 2) held-in evaluation where we train the released model on 2D-image-action-language pairs (\ie, 11 datasets selected from Open-X and RoboVQA dataset).
For the localization task, we compare with 2D grounding MLLM, namely Kosmos-2~\cite{peng2023kosmos} and CoVLM~\cite{li2023covlm}. Specifically, we use these models to detect 2D bounding boxes in a zero-shot manner and then transfer them to 3D bounding boxes using depth projection.

\textbf{Result analysis.}
In Tables~\ref{tab:pretrain_held_in}, 3D-VLA outperforms all 2D VLM methods on language reasoning tasks. We attribute it to the leverage of 3D information, which provides more accurate spatial information for reasoning. Besides, since our dataset contains a bunch of 3D localization annotations, 3D-VLA learns to localize the relevant objects, which helps the model focus more on key objects for reasoning.
Moreover, we find that 3D-LLM performs poorly on these robotic reasoning tasks, which demonstrates the necessity of collecting and training on a robotics-related 3D dataset.
In Table~\ref{tab:localization}, 3D-VLA demonstrates a marked superiority over the 2D baseline methods in terms of localization performance. This finding serves as compelling evidence of the efficacy of our annotation process, which supplies a substantial quantity of 3D annotations, thereby facilitating the acquisition of robust 3D localization capabilities within our model.

\begin{figure*}[t]
    \begin{center}
        \centerline{
        \includegraphics[width=0.90\linewidth]{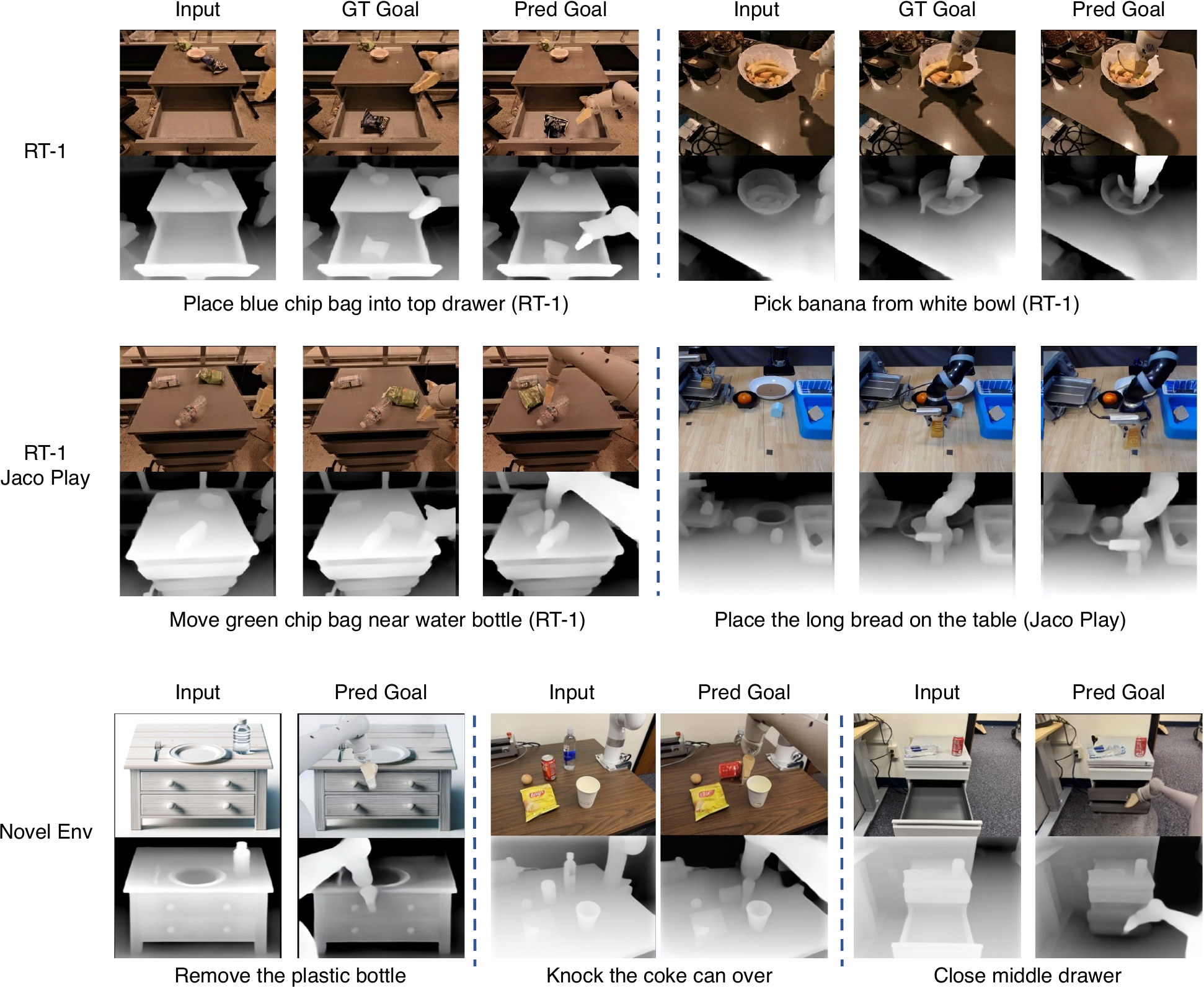}
        }
        \vspace{-2mm}
        \caption{Visualization of generated RGB-D goal images. The results in the first row are sampled from the test set of held-in training data while the second row is the unseen environments gathered from the Internet or daily life.}
        \label{fig:goal_image}
        \vspace{-3mm}
    \end{center}
\end{figure*}

\subsection{Multi-modal Goal Generation}
\textbf{Tasks.} We quantitatively evaluate the RGB goal and point cloud goal generation capability of 3D-VLA on Open-X test sets. We randomly sample 4000 episodes from the Open-X test set which 3D-VLA does not see in the training process. 

\textbf{Baselines.}
For image generation, we compare 3D-VLA with three types of image generation methods: 1) image-editing methods Instruct-P2P~\cite{brooks2023instructpix2pix}; 2) goal image/video generation methods SuSIE~\cite{black2023zeroshot}; 3) LLMs with image generation ability NeXT-GPT~\cite{wu2023next}.
For point cloud generation, we compare with text-to-3D diffusion model Point-E~\cite{nichol2022point}.

\textbf{Qualitative results.}
The image goal generation results are shown in Table~\ref{tab:goal_image1}. 
When compared with the existing generation methods that directly zero-shot transfers to the robotics domain (rows 1, 2, 3 in Table~\ref{tab:goal_image1}), 3D-VLA achieves a promising performance in terms of most metrics. This underscores the importance of training a world model using datasets specifically designed for robotics applications.
Even in a direct comparison with Instruct-P2P*, which was trained on the same robotics datasets we employed (row 4 in the table), 3D-VLA consistently outperforms it. This highlights that the integration of a large language model into 3D-VLA results in a more comprehensive and insightful comprehension of robotics manipulation instructions, leading to better goal image generation performance.
Furthermore, when we exclude the predicted bounding box from the input prompt (row 5), we observe a slight decrease in performance. This observation confirms the effectiveness of using these intermediate predicted bounding boxes as they assist the model in comprehending the overall scene, allowing the model to allocate more attention to the specific object mentioned in the given instruction, ultimately enhancing its ability to imagine the final goal images.

The point cloud generation results are presented in Table~\ref{tab:goal_pcd}.
3D-VLA with intermediate predicted bounding boxes performs the best. This outcome reinforces the significance of incorporating large language models and precise object localization in the context of comprehending both the instruction and the scene.

\textbf{Quantitative results.}
In the first row of Figure~\ref{fig:goal_image}, we visualize the generated RGB-D goal images on the test set of RT-1~\cite{brohan2022rt} and Jaco Play~\cite{dass2023jacoplay} datasets. These samples are not seen in the training process. Given the initial scenes and instructions, the 3D-VLA model consistently exhibits the capability to maintain the background elements unchanged while accurately identifying the target object of interaction and correctly modifying the states of these identified objects following the provided instructions.
The generated RGB-D goal images closely align both in terms of visual appearance and semantic content with the ground truth goal.
In addition to our controlled experimental settings, we extended our testing to encompass scenes captured from the internet or everyday life. In these diverse and uncontrolled environments, our 3D-VLA model consistently and robustly demonstrated its efficacy. 

\subsection{Embodied Action Planning}
\textbf{Tasks}
We evaluate the ability of \modelname for robot arm action prediction on two benchmarks, namely RLBench~\cite{james2020rlbench} and CALVIN~\cite{mees2022calvin}. 
We select three tasks from RLBench for evaluation. Besides, we also select var1 from the pick-up-cup task as an unseen task to test the model's generalization ability.
For CALVIN, we evaluate our model under the long-horizon multi-task language control setting, where the agent is required to execute 5 tasks sequentially. We train the agent on scenes A, B, C, D and test on scene D. 

\begin{table}[t]
\centering
\resizebox{\linewidth}{!}{
\begin{tabular}{l|cccc}
\hline
                          & Put         & Take     & Pick up & Pick up      \\
                          & Knife       & Umbrella & Cup     & Cup (unseen) \\ \hline
LanCon-Learn              & 28.8        & 45.6     & 23.2    & -            \\
LanCon-Learn w/ His.      & 32.2        & 50.8     & \textbf{44.2}    & -            \\
\modelname                & \textbf{68} & \textbf{52}       & 40      & 24           \\ \hline
\end{tabular}}
\caption{Evaluation of action planning on RLBench dataset.}
\label{tab:rlbench}
\vspace{-1mm}
\end{table}

\textbf{Baselines.}
For RLBench, we compare our model \modelname with LanCon-Learn~\cite{silva2021lancon}, which is a multi-task approach that can predict actions based on instruction-conditioned inputs. 
For CALVIN, we compare with MCIL~\cite{lynch2020language}, which is a conditional sequence-to-sequence variational autoencoder.

\textbf{Result analysis.}
As shown in Table~\ref{tab:rlbench}, \modelname surpasses or matches the baseline performance in most tasks within the RLBench action prediction, showing its planning capability. It's worth noting that the baseline uses history observations, object states, and current state information, whereas we only execute via open-loop control. Additionally, our generalization capability is proven in the pick-up-cup task.
In Table~\ref{tab:calvin}, \modelname also achieves promising results in CALVIN. We attribute the superiority to the ability to localize the objects of interest and imagine the goal state, which provides rich information for inferring actions.

\begin{table}[t]
\centering

\begin{tabular}{l|lllll}
\hline

       & \multicolumn{5}{c}{Tasks completed in a row}                                                                          \\
       & \multicolumn{1}{c}{1} & \multicolumn{1}{c}{2} & \multicolumn{1}{c}{3} & \multicolumn{1}{c}{4} & \multicolumn{1}{c}{5} \\ \hline
MCIL & 28.2                & 2.5                 & 0.3                 & 0                  & 0                   \\
3D-VLA   & \textbf{44.7}               & \textbf{16.3}                & \textbf{8.1}                 & \textbf{1.6}                   & 0                   \\ \hline
\end{tabular}
\caption{Evaluation of action planning on CALVIN dataset.}
\vspace{-3mm}
\label{tab:calvin}
\end{table}

\section{Conclusion}
In this paper, we introduce \modelname, a generative world model that can reason, understand, generate, and plan in the embodied environment.
We devise a novel data generation pipeline to construct a dataset including 2M 3D-Language-action data pairs to train our model. These data enable it to perform diverse tasks such as task caption, localization, goal image/point cloud generation, action prediction, etc. Our model uses 3D-LLM as the backbone and introduces interaction tokens to interact with the environment. We train a image to image and point to point diffusion model for embodied AI. They are further aligned by a projector with the LLM to enhance the LLM's multimodal generation capabilities. The experiment further shows that our \modelname has stronger capabilities in embodied tasks than the 2D baseline.

\section*{Impact Statement}
This paper introduces research aimed at pushing the boundaries of Machine Learning in the realm of robot manipulation. Given that robots operate in the physical world, the potential for collisions with objects and humans arises when the robot system is not adequately configured. To mitigate this issue, our approach involves initial training in a simulator environment followed by real-world deployment under human supervision, to minimize any adverse impacts.

\bibliography{example_paper}
\bibliographystyle{icml2024}

\newpage
\appendix
\onecolumn
\section{Model Implementation Details}

We use pretrained BLIP-2 FlanT5 as backbone. In the pretrain stage, we train 3D-VLAs for 30 epochs on $6\times32$ V100s, and validate every epoch. The batch size is set to 4 on each node during training. Additionally, we apply a linear warmup of the learning rate during the initial 1K steps, increasing from $10^{-8}$ to $10^{-5}$, followed by a cosine decay with a minimum learning rate of $10^{-6}$. In the alignment stage, we train 3D-VLAs for a maximum of epochs of 20 on $6\times64$ V100s. The batch size is set to 2 on each node for training. The AdamW optimizer is used, with $beta_1=0.9,beta_2=0.999$, and a weight decay of $0.05$. We use Distributed Data Parallel to train our models.

\section{Datasets Details}
\subsection{Details on Question Templates}
In this section, we show the question templates for data generation in Table~\ref{tab:ques_temp}.
We designed corresponding templates for six tasks. We design the templates for six tasks, and we replace the \texttt{INSTRUCTION}, \texttt{OBJECT}, \texttt{LOCATION}, and \texttt{ACTION} in each template with the information processed from each sample.

\begin{table}[H]
    \centering\resizebox{\linewidth}{!}{
        \begin{tabular}{ll}
            \toprule
            Tasks             & Templates                                                                                                                                                                        \\
            \midrule

            Verification      & The initial scene is \texttt{<}scene\texttt{>}\texttt{<}/scene\texttt{>} and the current scene is \texttt{<}scene\texttt{>}\texttt{<}/scene\texttt{>}.                           \\
                              & Instruction: \texttt{INSTRUCTION}. Finished? Answer: [yes/no]                                                                                                                    \\
            \midrule
            Task Caption      & The initial scene is \texttt{<}scene\texttt{>}\texttt{<}/scene\texttt{>} and the final scene is \texttt{<}scene\texttt{>}\texttt{<}/scene\texttt{>}.                             \\
                              & Describe the task. Answer: \texttt{INSTRUCTION}.                                                                                                                                 \\
            \midrule
            Localization      & The scene is \texttt{<}scene\texttt{>}\texttt{<}/scene\texttt{>}. Locate: \texttt{OBJECT}. Answer: \texttt{LOCATION}                                                             \\
            \midrule
            Dense Caption     & The scene is \texttt{<}scene\texttt{>}\texttt{<}/scene\texttt{>}. What is located at \texttt{LOCATION}? Answer: \texttt{OBJECT}                                                  \\
            \midrule
            Image or Point    & The initial scene is \texttt{<}scene\texttt{>}\texttt{<}/scene\texttt{>}. Instruction: \texttt{INSTRUCTION}.                                                                     \\
            Cloud Generation  & Generate the goal image (or point cloud). Answer: \texttt{<}image\texttt{>} (\texttt{<}pcd\texttt{>}) \texttt{INSTRUCTION} \texttt{<}/image\texttt{>} (\texttt{<}/pcd\texttt{>}) \\
            \midrule
            Action Prediction & \texttt{<}scene\texttt{>}\texttt{<}/scene\texttt{>}. \texttt{INSTRUCTION}. Predict \texttt{\{key/dense\}} actions. Answer: \texttt{ACTION}.                                      \\
            \bottomrule
        \end{tabular}}
    \caption{{Detailed on Question Templates.}}
    \label{tab:ques_temp}
\end{table}

\subsection{Details on ChatGPT-based Prompting}
\begin{figure}[ht]
    \centering
    \includegraphics[width=0.9\linewidth]{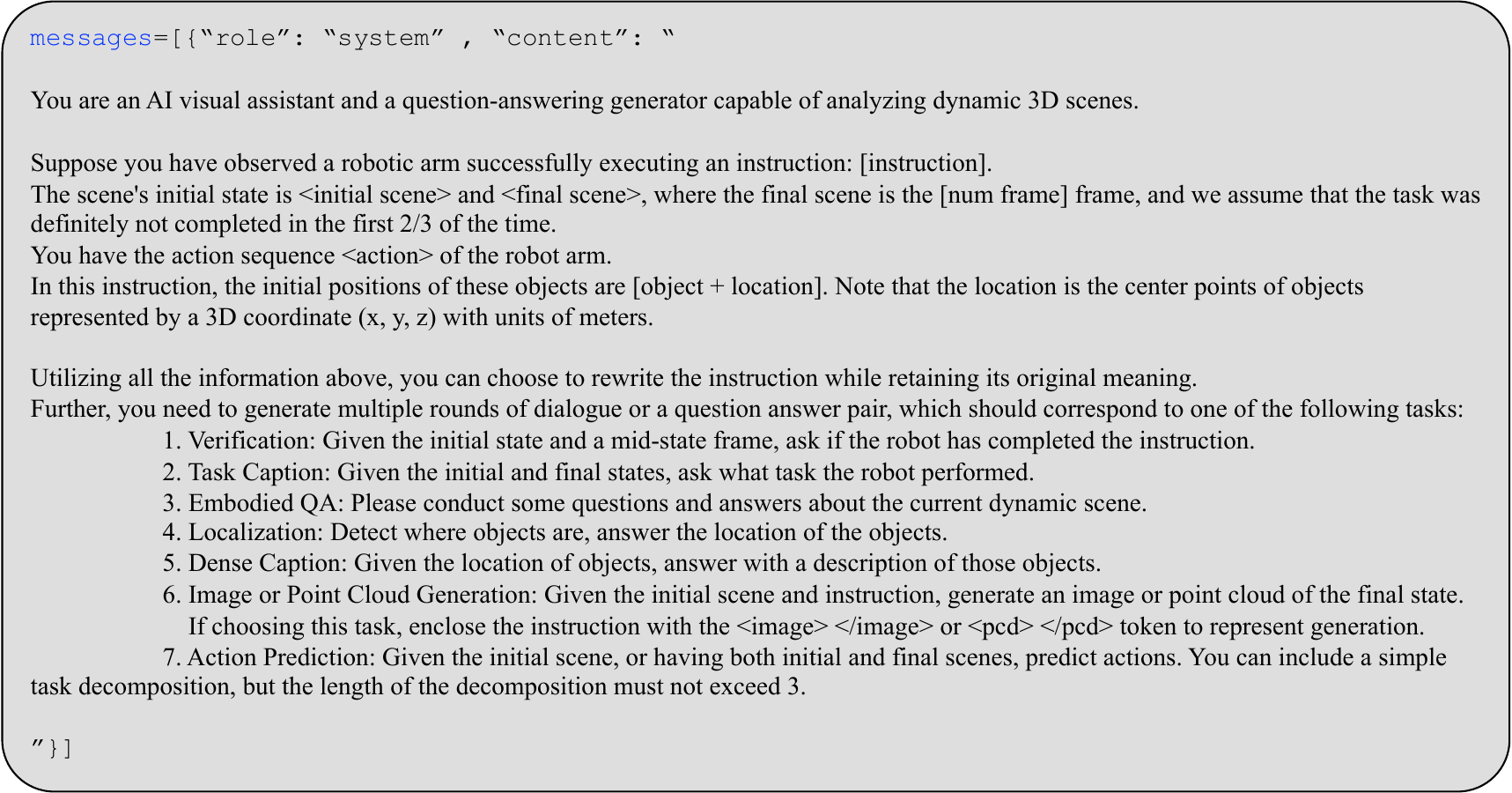}
    \caption{Prompt for ChatGPT-based data generation.}
    \label{fig:prompt-app}
\end{figure}

In this section, we show the prompt used in ChatGPT-based data generation in Figure~\ref{fig:prompt-app}. The ChatGPT version used in our paper is GPT-3.5-turbo-0125. We generate data for all seven tasks, and we provide all the information in the form of text, such as the instructions performed by the robot, total execution time, objects and their locations in the scene, etc. Additionally, for each prompt, we provide two manually written samples as guidance to direct ChatGPT towards more natural data generation.

\subsection{Details on Dataset Construction}
We show the number of the episodes and how we use them in table Table~\ref{dataset_overview}. We utilize two main categories of datasets, namely robotics datasets and human object interaction (HOI) datasets. For the former, we filtered out complex scene datasets to prevent the Grounded-SAM from detecting incorrect object locations. However, within the same robotics dataset, the background settings are largely the same. Therefore, in the Goal Generation tasks, we included HOI datasets to better allow the diffusion model to learn diverse scenes, object interaction methods, etc.

{\renewcommand{\arraystretch}{1.3}
\begin{sidewaystable}[]
  \resizebox{\textwidth}{!}{
    \begin{tabular}{l|c|ccccc|ccc|c}
      \hline
      \multicolumn{1}{c|}{}                                                                            &
      \multicolumn{1}{c|}{}                                                                            &
      \multicolumn{5}{c|}{Reasoning and Perception}                                                    &
      \multicolumn{3}{c|}{Goal Generation}                                                             &
      \multicolumn{1}{c}{Decision Making}                                                                                                                                                                                          \\
      \multicolumn{1}{c|}{\multirow{-2}{*}{Dataset}}                                                   &
      \multicolumn{1}{c|}{\multirow{-2}{*}{\# of Used Episodes}}                                       &
      \multicolumn{1}{c}{\begin{tabular}[c]{@{}c@{}}Embodied QA\\ What-if QA\end{tabular}}             &
      \multicolumn{1}{c}{\begin{tabular}[c]{@{}c@{}}Task Caption\\ (w/ Object Grounding)\end{tabular}} &
      \multicolumn{1}{c}{Dense Caption}                                                                &
      \multicolumn{1}{c}{Verification}                                                                 &
      \multicolumn{1}{c|}{Detection}                                                                   &
      \multicolumn{1}{c}{Image}                                                                        &
      \multicolumn{1}{c}{Depth}                                                                        &
      \multicolumn{1}{c|}{Point Cloud}                                                                 &
      \multicolumn{1}{c}{Action Prediction}
      \\ \hline
      \rowcolor[HTML]{C0C0C0}
      Robotics Datasets                                                                                & 305k & \checkmark & \checkmark & \checkmark & \checkmark & \checkmark & \checkmark & \checkmark & \checkmark & \checkmark \\
      BC-Z                                                                                             & 40k  & \checkmark & \checkmark & \checkmark & \checkmark & \checkmark & \checkmark & \checkmark & \checkmark & \checkmark \\
      Bridge                                                                                           & 25k  & \checkmark & \checkmark & \checkmark & \checkmark & \checkmark & \checkmark & \checkmark & \checkmark & \checkmark \\
      CALVIN                                                                                           & 10k  & -          & -          & -          & \checkmark & \checkmark & \checkmark & \checkmark & \checkmark & \checkmark \\
      Dobb-E                                                                                           & 20k  & \checkmark & \checkmark & -          & \checkmark & -          & \checkmark & \checkmark & \checkmark & \checkmark \\
      Fractal                                                                                          & 70k  & \checkmark & \checkmark & \checkmark & \checkmark & \checkmark & \checkmark & \checkmark & \checkmark & \checkmark \\
      Jaco Play                                                                                        & 0.9k & \checkmark & \checkmark & \checkmark & \checkmark & \checkmark & \checkmark & \checkmark & \checkmark & \checkmark \\
      Lang Table                                                                                       & 13k  & \checkmark & \checkmark & -          & -          & -          & \checkmark & \checkmark & \checkmark & \checkmark \\
      Mutex                                                                                            & 1.5k & \checkmark & \checkmark & \checkmark & \checkmark & \checkmark & \checkmark & \checkmark & \checkmark & \checkmark \\
      Pick\&Place                                                                                      & 1.3k & \checkmark & \checkmark & \checkmark & \checkmark & \checkmark & \checkmark & \checkmark & \checkmark & \checkmark \\
      Play Fusion                                                                                      & 0.5k & \checkmark & \checkmark & \checkmark & \checkmark & \checkmark & \checkmark & \checkmark & \checkmark & \checkmark \\
      Playing Food                                                                                     & 4.2k & \checkmark & \checkmark & -          & \checkmark & -          & \checkmark & \checkmark & \checkmark & \checkmark \\
      RH20T                                                                                            & 2.0k & \checkmark & \checkmark & \checkmark & \checkmark & \checkmark & \checkmark & \checkmark & \checkmark & \checkmark \\
      RLBench                                                                                          & 50k  & -          & -          & -          & \checkmark & \checkmark & \checkmark & \checkmark & \checkmark & \checkmark \\
      Roboturk                                                                                         & 2.0k & -          & -          & -          & \checkmark & -          & \checkmark & \checkmark & \checkmark & \checkmark \\
      RoboVQA                                                                                          & 61k  & \checkmark & -          & -          & -          & -          & -          & -          & -          & -          \\
      Taco Play                                                                                        & 3.2k & \checkmark & \checkmark & \checkmark & \checkmark & \checkmark & \checkmark & \checkmark & \checkmark & \checkmark \\
      \hline
      \rowcolor[HTML]{C0C0C0}
      HOI Datasets                                                                                     & 11k  & -          & -          & -          & -          & -          & \checkmark & \checkmark & \checkmark & -          \\
      Epic Kitchen                                                                                     & 6k   & -          & -          & -          & -          & -          & \checkmark & \checkmark & \checkmark & -          \\
      HOI4D                                                                                            & 5k   & -          & -          & -          & -          & -          & \checkmark & \checkmark & \checkmark & -          \\
      \hline
      \rowcolor[HTML]{EFEFEF}
      All Datasets                                                                                     & 316k & \checkmark & \checkmark & \checkmark & \checkmark & \checkmark & \checkmark & \checkmark & \checkmark & \checkmark \\ \hline
    \end{tabular}
  }
  \caption{Datasets used in our paper. We categorize them into four categories: Robotics, HOI, and Room datasets.}
  \label{dataset_overview}
\end{sidewaystable}}

\section{More Visualization Results about Goal Generation}
\begin{figure*}[t]
    \begin{center}
        \centerline{
            \includegraphics[width=0.90\linewidth]{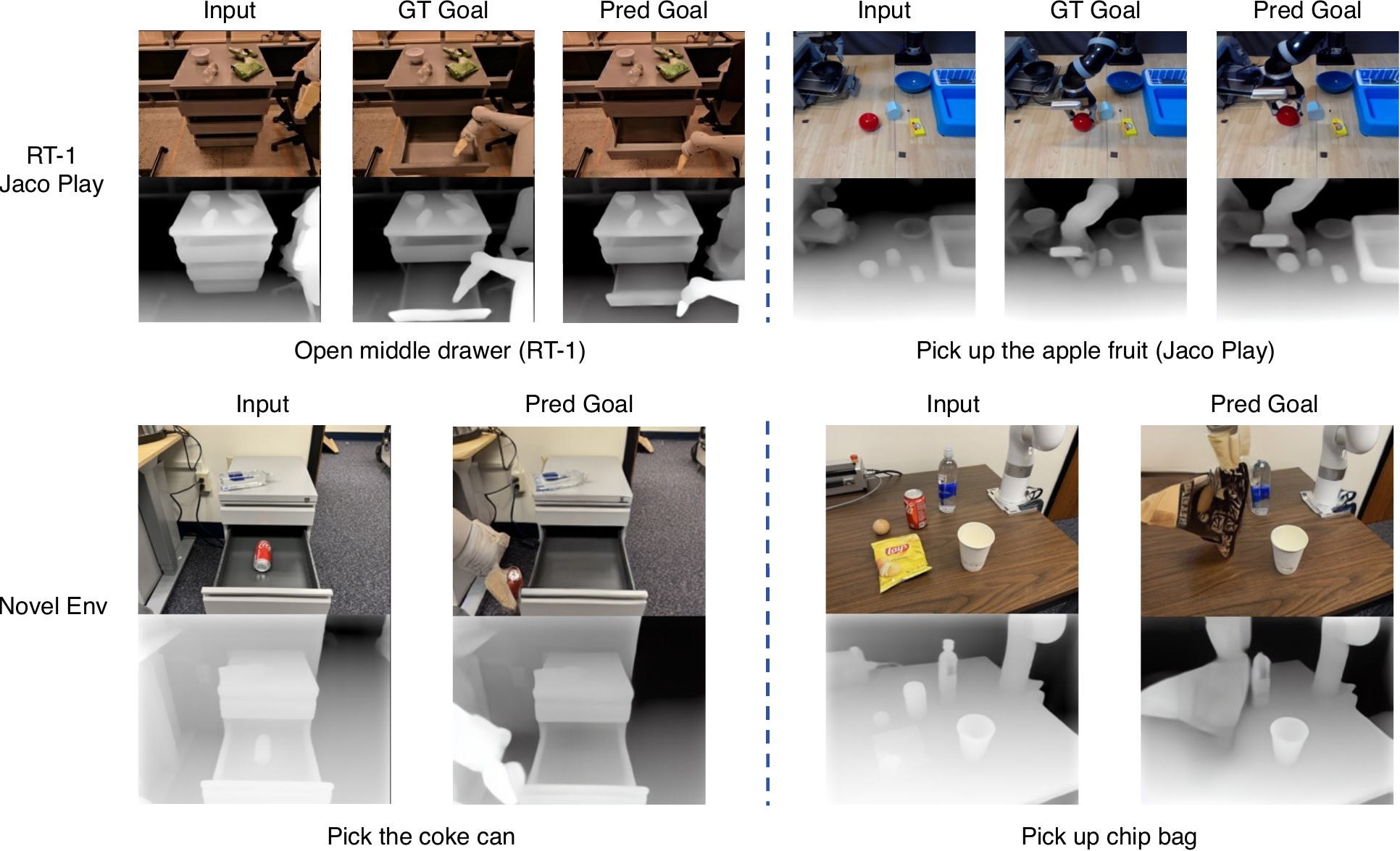}
        }
        \caption{Visualization of generated RGBD goal images. The results in the first row are sampled from the test set of held-in training data while the second row are the unseen environments gathered from daily life.}
        \label{fig:goal_app}
    \end{center}
\end{figure*}

\begin{figure*}[t]
    \begin{center}
        \centerline{
            \includegraphics[width=0.90\linewidth]{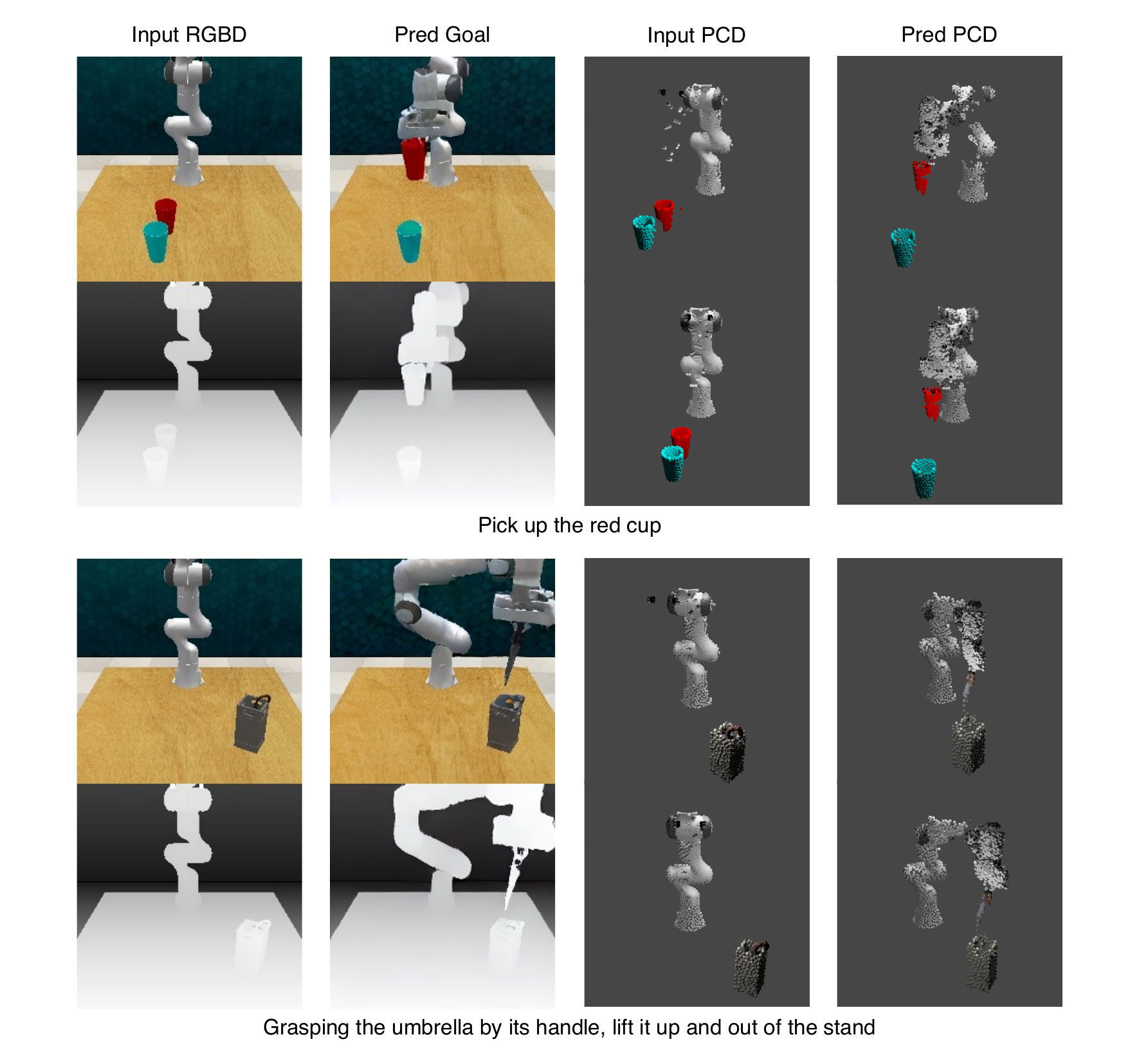}
        }
        \vspace{-2mm}
        \caption{Visualization of generated RGB-D goal images and goal point cloud. (RLBench)}
        \label{fig:goal_rlbench}
        \vspace{-3mm}
    \end{center}
\end{figure*}

We show more qualitative examples in Figure~\ref{fig:goal_app}, \ref{fig:goal_rlbench}.

\end{document}